\documentclass{article}
\usepackage{times}
\usepackage{epsfig}
\usepackage{graphicx}
\usepackage{amsmath}
\usepackage{amssymb}
\usepackage{booktabs}
\usepackage{bbding}
\usepackage{caption}
\usepackage{subcaption}
\usepackage{array}
\usepackage{multirow}
\usepackage{subfloat}
\usepackage{float}
\usepackage{color}
\usepackage{comment}
\usepackage{colortbl}
\usepackage{pifont} 

\usepackage{enumitem} 
\def\x{$\times$}

\newcolumntype{x}[1]{>{\centering\arraybackslash}p{#1pt}}
\newcolumntype{y}[1]{>{\raggedright\arraybackslash}p{#1pt}}
\newcolumntype{z}[1]{>{\raggedleft\arraybackslash}p{#1pt}}
\newcolumntype{k}[1]{>{\raggedright\arraybackslash}p{#1pt}}

\newlength\savewidth\newcommand\shline{\noalign{\global\savewidth\arrayrulewidth\global\arrayrulewidth 1pt}\hline\noalign{\global\arrayrulewidth\savewidth}}

\definecolor{xycolor}{RGB}{60, 120, 216}

\definecolor{wcolor}{RGB}{103, 78, 167}

\definecolor{dcolor}{RGB}{166, 77,21}

\definecolor{gcolor}{RGB}{204, 102, 153}

\definecolor{tcolor}{RGB}{34,139,34}

\definecolor{iterc}{RGB}{91,196,159}

\definecolor{epochc}{RGB}{96,172,252}

\definecolor{eicolor}{RGB}{153, 51, 102}

\definecolor{orange}{RGB}{237, 125, 49}

\usepackage[pagebackref=false,breaklinks=true,colorlinks,urlcolor=blue,citecolor=blue,linkcolor=blue,bookmarks=false]{hyperref}

\definecolor{mygray}{gray}{0.92}
\definecolor{baselinecolor}{gray}{.9}

\usepackage[capitalize]{cleveref}
\crefname{section}{Sec.}{Secs.}
\Crefname{section}{Section}{Sections}
\Crefname{table}{Table}{Tables}
\crefname{table}{Tab.}{Tabs.}
\usepackage[accsupp]{axessibility}

\usepackage{spconf,amsmath,epsfig}
\usepackage{indentfirst} 
\usepackage[ruled,linesnumbered]{algorithm2e}
\let\OLDthebibliography\thebibliography
\renewcommand\thebibliography[1]{
  \OLDthebibliography{#1}
  \setlength{\parskip}{0pt}
  \setlength{\itemsep}{0pt plus 0.3ex}
}
\usepackage{biblatex}
\usepackage{amssymb}
\makeatletter

\newcommand{\Rmnum}[1]{\expandafter\@slowromancap\romannumeral #1@}
\makeatother

\begin{document}\sloppy

\def\x{{\mathbf x}}
\def\L{{\cal L}}

\title{SFMViT : SlowFast Meet ViT in Chaotic World}
%
\name{Jiaying Lin$^*$$^{12}$, Jiajun Wen$^*$$^{12}$, Mengyuan Liu$^\dagger$$^1$, Yue Li$^2$, Jinfu Liu$^2$, Baiqiao Yin$^2$}
\vspace{-2mm}
\address{
$^1$National Key Laboratory of General Artificial Intelligence, Peking University, Shenzhen Graduate School\\
$^2$School of Intelligent Systems Engineering, Sun Yat-sen University \\
$^*$ means co-first authors with equal contributions. \\
The corresponding author is Mengyuan Liu (e-mail: liumengyuan@pku.edu.cn). }


\maketitle

\begin{abstract}
The task of spatiotemporal action localization in chaotic scenes is a challenging task toward advanced video understanding. Paving the way with high-quality video feature extraction and enhancing the precision of detector-predicted anchors can effectively improve model performance. To this end, we propose a high-performance dual-stream spatiotemporal feature extraction network SFMViT with an anchor pruning strategy. The backbone of our SFMViT is composed of ViT and SlowFast with prior knowledge of spatiotemporal action localization, which fully utilizes ViT's excellent global feature extraction capabilities and SlowFast's spatiotemporal sequence modeling capabilities. Secondly, we introduce the confidence maximum heap to prune the anchors detected in each frame of the picture to filter out the effective anchors. These designs enable our SFMViT to achieve a mAP of 26.62\% in the Chaotic World dataset, far exceeding existing models.
Code is available at \textcolor{blue}{https://github.com/jfightyr/SlowFast-Meet-ViT}.
\end{abstract}
\section{Introduction}
{Spatiotemporal} action localization, which involves localizing persons and recognizing their actions from videos, is an important task that has gained attention in recent years \cite{gu2018ava, pan2021actor, feichtenhofer2019slowfast, chen2023efficient, ryali2023hiera,liu2022generalized}. Spatiotemporal feature modeling refers to learning the temporal and spatial characteristics present in videos as the most fundamental and critical part of action recognition tasks\cite{xie2018rethinking}. However, the subpar quality of feature extraction by current methods hampers the advancement of this field, leaving room for improvement in model performance.

The Chaotic World \cite{Ong_2023_ICCV}, a newly proposed dataset \cite{kay2017kinetics, sigurdsson2016hollywood,gu2018ava} in this task, presents more complex and chaotic scenes compared to previous datasets, thus imposing greater demands on comprehending intricate scenes and interactions among individuals. It's not uncommon to encounter difficulties when applying traditional video feature extraction modules such as I3D \cite{carreira2017quo} or SlowFast \cite{feichtenhofer2019slowfast} directly to datasets like Chaotic World. 
The complexity and chaotic nature of scenes in the Chaotic World dataset may necessitate more advanced or customized approaches for feature extraction and modeling to attain satisfactory results.

On the one hand, VideoMAEv2 \cite{wang2023videomae} has achieved state-of-the-art performance across multiple downstream tasks by employing a straightforward yet effective video self-supervised pretraining paradigm. It is reasonable to expect that robust video feature extraction networks pre-trained on large-scale video datasets will perform effectively.
SlowFast \cite{feichtenhofer2019slowfast}, a standard backbone network in the Video Recognition field, excels at capturing action features with strong temporal correlations, but it may be sensitive to disruptions in complex environments, resulting in reduced robustness across different scenes. Conversely, Vision Transformer \cite{dosovitskiy2020image}, trained on extensive video data, demonstrates robustness in spatiotemporal feature modeling, capable of capturing diverse spatiotemporal features in complex scenes. 

Considering the strengths of both models, we introduce the dual-stream \textbf{SFMViT} video spatiotemporal modeling network. This network utilizes the advantages of SlowFast and ViT to enhance the robustness and effectiveness of spatiotemporal feature modeling in video recognition tasks.
In addition, the action detection of the Chaotic World dataset needs to focus on the action recognition of specific people, so effective actor detection is also very critical. To this end, we designed a \textbf{Confidence Pruning Strategy} to avoid redundant actor extracted by Yolo, thereby ensuring that ACAR \cite{pan2021actor} works better and improving the generalization and accuracy of the SFMViT model.

We conduct extensive experiments on the challenging Chaotic World dataset for spatiotemporal action localization. Our proposed SFMViT leads to significant improvements in localizing spacial-temporal actions. Our contributions are summarized as three-fold:
\vspace{-1mm}
\noindent
\begin{itemize}[leftmargin=*]
\item We have introduced the dual-stream spatiotemporal feature modeling network SFMViT, which integrates SlowFast's ability to capture temporal features with Vision Transformer's capability in complex scene spatiotemporal modeling. This fusion enhances overall spatiotemporal modeling capabilities in complex scenes.
\vspace{-2mm}
\item We introduce a Confidence Pruning Strategy to find the most suitable anchor number of the instances, which can be used to prune anchors and filter out the optimal anchors while taking into account the efficiency and performance, increasing the mAP by nearly 2\%.
\vspace{-2mm}
\item We achieve SOTA performances with significant margins on the Chaotic World datasets. At the time of submission, our method ranks first on the 2024 MMVRAC leaderboard. 
\end{itemize}

\vspace{-4mm}
\section{Related Work}

\subsection{Spatiotemporal Action Localization}
\vspace{-2mm}
The spatiotemporal action localization problem is typically composed of a detector for participant localization on keyframes and a 3D backbone for video feature extraction. Currently, it can be mainly classified into two types based on whether the detector is embedded within the backbone model. One approach implements localization and feature classification as two independent backbones in a two-stage pipeline, while the other employs a unified backbone for spatiotemporal localization and feature classification. Unified backbone models often face challenges such as high complexity, difficulty in optimization, and high training costs \cite{chen2021watch,zhao2022tuber,wu2023stmixer,chen2023efficient}. Therefore, the most advanced methods currently \cite{feichtenhofer2020x3d,pan2021actor,tang2020asynchronous,wang2023videomae} tend to adopt a two-stage pipeline model.
In our model, considering that unified backbone models incur higher training costs compared to two-stage pipeline models for achieving the same performance, we select the two-stage pipeline model as the baseline.

\vspace{-4mm}
\subsection{Video Foundation Models}
\vspace{-2mm}
Video Foundation Models (ViFMs) have emerged as a pivotal area of research in artificial intelligence \cite{li2023unmasked,wang2024internvideo2,wang2023videomae}, with a focus on constructing models capable of comprehending and generating video content through extensive learning from video data. These models have exhibited remarkable performance across various downstream tasks, including action recognition, video-text retrieval, video question answering, and video generation.

InternVideo2 \cite{wang2024internvideo2} represents a recent advancement achieved through progressive training paradigms, demonstrating optimal performance in action recognition, video-text tasks, and video-centric dialogues. Its integration of diverse self-supervised learning frameworks highlights the potential for enhancing the performance of video foundation models. VideoMAEv2 \cite{wang2023videomae} has trained a more generalized video foundation model capable of adapting to various downstream tasks by scaling and diversifying pre-training video data. 
These models and methods play a crucial role in advancing video foundation model research and applications, greatly enhancing the model's performance in capturing video features.

\vspace{-2mm}

\section{Method}
\begin{figure*}[t]
\centering
	\includegraphics[width=0.95\linewidth]{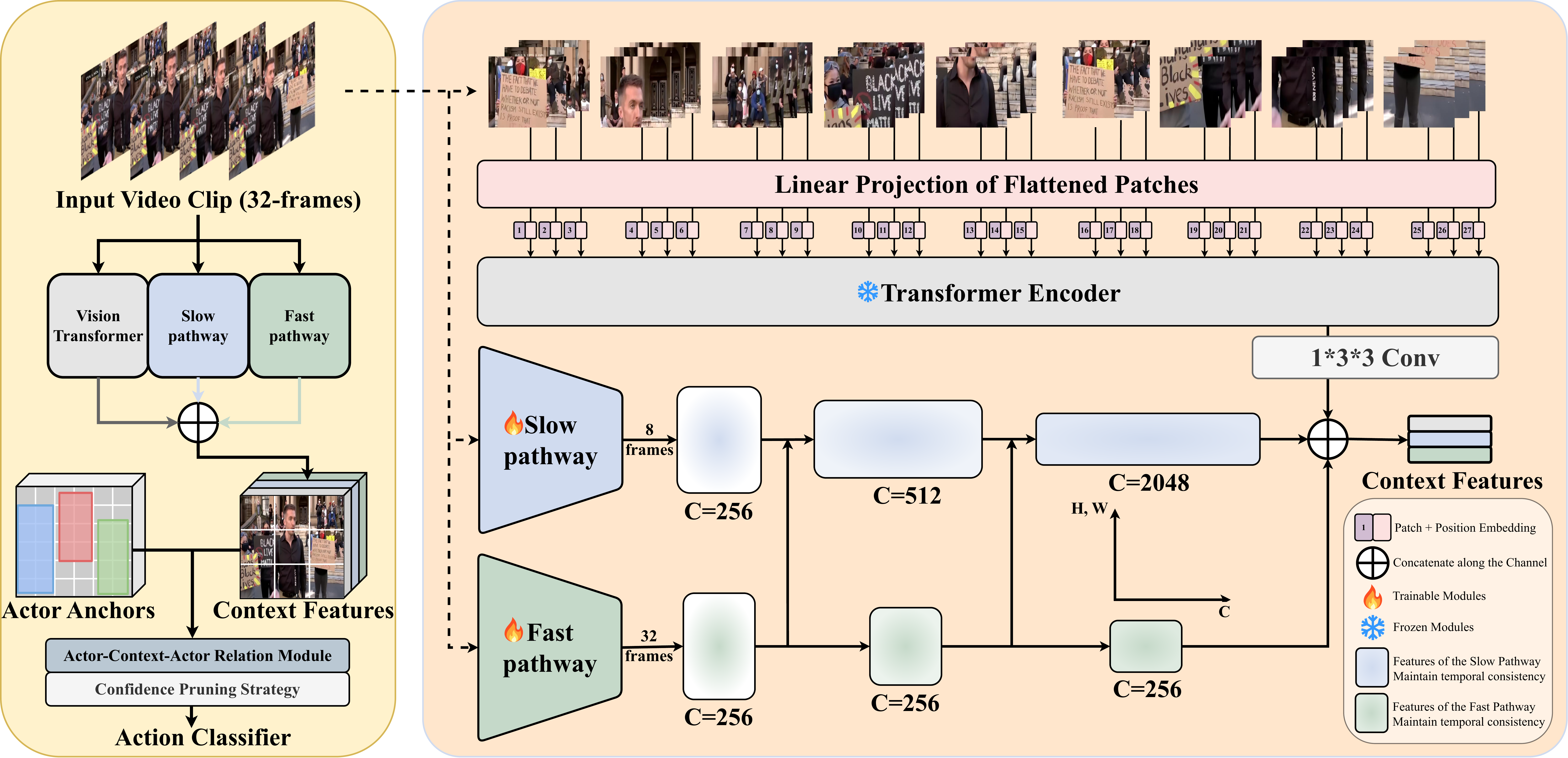}
        \caption{\textbf{Framework of our proposed SFMViT}. The left yellow background box diagram shows the architecture flow of our entire {spatiotemporal action localization} task, and the right orange background box diagram shows the details of the SFMViT module we propose. The 32-frame input video clip goes through the ViT and SlowFast dual-stream networks to obtain spatiotemporal context features. {The coordinate axes illustrate the changes in the spatial size and channel dimensions of the feature maps in the SlowFast branch.} {Anchors} are obtained by YOLO series object detectors based on the detection results at keyframes to get individual features from context features. These features of the target individuals go through high-order relation reasoning by the ACAR module \cite{pan2021actor} and our proposed Confidence Pruning Strategy before the final action category classification.}
    \label{fig: SFMViT}
\end{figure*}

\subsection{Overall  Architecture}
\vspace{-2mm}
We first introduce our overall framework for action localization, where the proposed dual-stream spatiotemporal feature modeling network SFMViT is the key module as the backbone part. The framework is designed to detect all individuals in an input video clip and estimate their action labels. It is constructed using a pre-existing person detector (e.g. YOLOv9 \cite{wang2024yolov9} ) and our proposed video backbone network SFMViT.
The features of individuals and their surroundings are subsequently analyzed by the established ACAR module \cite{pan2021actor} , which incorporates a long-term Actor-Context Feature Bank for the final action prediction.

In detail, the person detector operates on the key frame of the input clip and obtains N detected actors. The detection boxes obtained from the person detector will be further used as annotations for our backbone network, guiding our final action classification of individuals. Simultaneously, our proposed visual feature extraction model SFMViT will extract global spatiotemporal features from input video clips. These feature maps will undergo average pooling along the temporal dimension to reduce computational costs, resulting in feature maps like $ X \in \mathbb{R}^{C \times H \times W} $, and C, H, W are channel, height and width respectively. Guided by annotations, these feature maps will further produce a series of N actor features, each containing spatiotemporal context features. Subsequently, these actor features will be fed into the ACAR module for higher-order relationship reasoning, culminating in the classification of the inferred features.

\vspace{-2mm}
\subsection{SFMViT Module}
\textbf{SFMViT.} Leveraging the inherent advantages of the aforementioned architectures, we have adopted the ViT model as our video feature extractor. Additionally, we have incorporated SlowFast as an auxiliary network with domain knowledge in Video Recognition. This integration has led to the development of our proposed SFMViT model architecture. More specifically, we obtain the fused features of the SlowFast and ViT models as:
\begin{equation}
f^{\text{SFMViT}}_{x,y} = [ f^{V}_{x,y}, f^{S}_{x,y}, f^{F}_{x,y} ]
\end{equation}
x,y represent the spatial location (x, y). [·,·,·] denotes concatenation along the channel dimension. $f^S$ and $f^F$ refer to feature maps obtained from {Slow pathway and Fast pathway respectively} and both of them have already undergone average pooling along the temporal dimension. Especially, $f^{V}$ obtained from the ViT model needs to undergo 3D convolution first to achieve the same spatial dimensions as $f^S$ and then undergo temporal average pooling. Finally, we can get a fused feature map like $ f^{SFMViT}_{x,y} \in \mathbb{R}^{C \times H \times W} $, where $C$ equals the sum of the channel numbers of the three original feature maps.

\textbf{Vision Transformer.} ViT employs a global attention mechanism by transforming videos into patches across spatial and temporal dimensions. Combined with the Transformer architecture, it can effectively captures the overall information available in videos. 
We select the pretrained ViT models on extensive datasets such as Kinetics700 \cite{kay2017kinetics} and AVA \cite{gu2018ava} to form a robust foundation for transfer learning across diverse downstream tasks.
\raggedbottom

\textbf{SlowFast Pathways.} The SlowFast feature extraction branch can capture strongly correlated temporal action signals and assist in extracting domain-specific features in our model. We follow the basic instantiation of SlowFast 8×8, R50 as stage $res_{5}$ defined in \cite{feichtenhofer2019slowfast}.
For the Fast pathway, it has a lower channel capacity, such that it can better capture motion while trading off model capacity. Spatial downsampling is performed with stride $2^2$  convolution in the center filter of the first residual block in each stage of both the Slow and Fast pathways.

\subsection{Confidence Pruning Strategy}
\vspace{-2mm}
It needs to be clarified that mAP is the mean average precision of each action category:
\begin{equation}
mAP = \frac{1}{N} \sum_{i=1}^{N} AP(a_i)
\end{equation}

N represents the total number of action categories
; $ a_i$ represents the i-th action class; and $ AP()$ denotes the average precision for a particular class:
\vspace{-2mm}
\begin{equation}
AP = \frac{1}{M} \sum_{k=1}^{k} Precision(k)
\end{equation}

\vspace{-2mm}
\noindent M denotes the number of positive samples in the test set, and $ Precision(s)$ denotes the precision for the top k test samples.

Because the task of spatiotemporal action localization focuses solely on the actions of key individuals, the detector, if left unprocessed, tends to generate a significant number of unnecessary anchors, leading to resource wastage. Previous methods typically involve extensive training of the detector on the target dataset, which requires substantial resources, takes a long time, and exhibits poor transferability. To address this problem, SFMViT adopts a strategy to store the predicted anchors of each $\{img\_id, time\}$ pair in the maximum heap based on the confidence score, and set the maximum number of nodes as the capacity, so that keep the top capacity anchors with the highest confidence.

It is worth noting that setting the capacity too high or too low can lead to performance degradation and resource wastage, as we will demonstrate in Section \ref{sec:capacity}. Furthermore, the optimal capacity value exhibits randomness across different datasets and models, making it inappropriate to fix a single capacity value, as done in prior approaches \cite{pan2021actor}. 
Our SFMViT employs a multi-threaded iterative approach 
(Algorithm \ref{alg:mAP_calculation})
, which empirically sets {the range for capacity enumeration (abbreviated as capacity\_range)} and quickly identifies the optimal capacity within this range. This approach prunes the predicted results to retain favorable anchors for the model, thereby capturing the optimal mAP value. Our strategy significantly improves model performance in a short time, aligning with the requirement for fast results in competitive scenarios.

\vspace{-3mm}
\begin{algorithm}
\footnotesize
    \caption{Confidence Pruning Strategy}
    \label{alg:mAP_calculation}
    \SetKwInOut{Input}{Input}
    \SetKwInOut{Output}{Output}
    \SetKwFunction{Insert}{Insert}
    \SetKwFunction{Replace}{Replace}
    \SetKwFunction{CalculateMAP}{CalculateMAP}

    \Input{Dataset, $\text{capacity\_range}$}
    \Output{mAP for each capacity value}

    \BlankLine
    \For{$\text{cap} \in \text{capacity\_range}$}{
        Initialize a tree structure with capacity = $\text{cap}$\;
        \ForEach{data point $\text{dp}$ \textbf{in} Dataset}{
            \If{number of entries in the tree $<$ cap}{
                \Insert{data point into the tree}\;
            }
            \ElseIf{score of data point $>$ score of least confident entry in the tree}{
                \Replace{least confident entry with current data point}\;
            }
        }
        \CalculateMAP{}\;
    }
\end{algorithm}

\vspace{-5mm}
\section{Experiments}
\label{experiment} 
\vspace{-2mm}
\subsection{Dataset}
\vspace{-2mm}
We evaluate our model on the Chaotic World dataset \cite{Ong_2023_ICCV}, a chaotic scene dataset containing multi-task annotations, with 299,923 annotated instances for spatiotemporal action localization, with 50 action categories, and with time ranges from 5 to 200 seconds per action (10 seconds on average). Officially, the data labels are divided into training sets and test sets according to AVA \cite{gu2018ava} format, and we experimented according to officially provided training and test sets.
\vspace{-2mm}
\subsection{Experimental Setup} 
\vspace{-2mm}

\textbf{Person Detector.} We used Yolov8 as our base object detector, which is pre-trained on AVA \cite{gu2018ava} and WiderPerson \cite{8764496}, then we fine-tuned it on our Chaotic World dataset. Due to different training strategies and varying confidence threshold settings, there was a significant impact on the results. Therefore, besides testing on the detector, we also conducted ablation experiments using Ground Truth anchors as the result of the first-stage object detection.

\textbf{Backbone Network.} We will use SFMViT proposed above as our video feature extractor. Specifically, we freeze the ViT model branch and set the SlowFast model branch to a trainable state.

\textbf{Training and inference.} All experiments were conducted 
on 4 * GeForce RTX 3090 GPUs. For the Chaotic World dataset, all given segments were cropped to the central 32 frames. Our model sets the capacity enumeration range to [50, 2200] and utilizes the stochastic gradient descent (SGD) optimizer for training, with Nesterov momentum of 0.9 and weight decay of \(10^{-7}\). We employed cross-entropy as the loss function. The training epochs and learning rate were set to 90 and 0.01, respectively, with a batch size of 3 per GPU for each training epoch.

\vspace{-3mm}
\subsection{Comparison with the state-of-the-art methods}
\vspace{-2mm}
We compared our SFMViT with state-of-the-art methods on the Chaotic World dataset, and SFMViT outperformed all previous models, as shown in Table \ref{table:all_res}. Currently, the best-performing model on the Chaotic World dataset is IntelliCare proposed in its paper, which is a multi-task model that leverages information exchange between tasks to aid performance. Our model achieves a performance improvement of 10.37\% over IntelliCare in the task of spatiotemporal action localization, even without additional information.
\begin{table}[t]
    \footnotesize
    \centering
    \setlength{\tabcolsep}{2pt}
    \begin{tabular}{x{40}x{80}x{80}x{35}} 
        \shline 
        \textbf{Model} & \textbf{Backbone} & \textbf{Pre-training Dataset} & \textbf{mAP} \\ 
        \shline 
        ACAR~\cite{pan2021actor} & SF-R50-NL & K400(ft:AVA v2.2) & 9.88 \\
        \rowcolor{baselinecolor} Hiera~\cite{ryali2023hiera} & Hiera-Large & K400 & 10.94 \\
        ACRN~\cite{sun2018actor} & I3D~\cite{carreira2017quo} & / & 11.91 \\
        \rowcolor{baselinecolor} ACRN~\cite{sun2018actor}& SlowFast~\cite{feichtenhofer2019slowfast}  & / & 12.90 \\
        SlowFast~\cite{feichtenhofer2019slowfast} & SF-R101-NL & K600 & 13.24 \\
        \rowcolor{baselinecolor} ACAR~\cite{pan2021actor} & SF-R101-NL & K600 & 13.99 \\
        IntelliCare~\cite{Ong_2023_ICCV} & I3D~\cite{carreira2017quo} & / & 16.25 \\
        \rowcolor{baselinecolor} $\mathbf{SFMViT}$$^\ddag$ & ViT-Giant+SF-R50 & K400(ft:AVA v2.2) & \textbf{26.62} \\
        \shline
    \end{tabular}
    \vspace{-2mm}
    \caption{\textbf{Comparison of the performance of SFMViT with other resolutions.} ``ft'' denotes fine-tuning on this dataset. {``/ '' represents unclear about its pre-training dataset. }}
    \label{table:all_res}
    \vspace{-4mm}
\end{table}
\setlength{\tabcolsep}{1.4pt}

\vspace{-2mm}
\subsection{Ablation Studies}
\vspace{-2mm}

\setlength{\tabcolsep}{2pt}
\begin{table*}[t]
\footnotesize
    \hspace{-2mm} 
    \begin{minipage}[t]{0.42\linewidth}
        \centering
        \begin{subtable}[t]{1.\linewidth}
            \centering
            \begin{tabular}{lll}
            \shline
            \textbf{Ablation object} & \textbf{Method} & \textbf{frame-mAP} \\
            \shline
            \textbf{Backbone} & w/o SF & 18.25 \\
            (capacity = 50) & w/o ViT & 9.88 \\
            & SFMViT & \textbf{24.67} \\
            \hline
            \textbf{Capacity} & cap=10 & 22.46 \\
            (backbone: SFMViT) & cap=50 & 24.67 \\
            & cap=500& 25.87 \\
            & \textbf{cap=1599(OURS)}  & \textbf{26.62} \\
            & cap=3000 & 22.18 \\
            \shline
            \end{tabular}
            \caption{\textbf{Comparison of frame-mAP under different ablation settings.}}
            \label{tab:ablation-comparison}
        \end{subtable}
        
    \end{minipage}
    \hspace{2mm} 
    \begin{minipage}[t]{0.5\linewidth}
        \centering
        \begin{subtable}[t]{1.\linewidth}
            \renewcommand{\arraystretch}{1.12} 
            \begin{tabular}{lcccc} 
                \shline \textbf{Backbone Type}&\textbf{Model} & \textbf{Backbone} & \textbf{Pre-training Dataset} & \textbf{mAP} \\
                \shline 
                &ACAR~\cite{pan2021actor} & SF-R50-NL & K400(ft:AVA v2.2) & 31.25\\
                \textbf{Single-stream} &Hiera~\cite{ryali2023hiera}& Hiera-Large & K400 & 32.88\\
                &ViT~\cite{dosovitskiy2020image} & ViT-Giant & K400 
                & 37.57 \\
                \hline &SlowFast+ViT& SF-R50-NL+ViT-Giant& K400(ft:AVA v2.2) & 29.54 \\
                \textbf{Multi-stream} &+Hiera$^\ddag$& +Hiera-Large& +K400 &  \\
                &Hiera+ViT$^\ddag$ & Hiera-Large+ViT-Giant & K400,AVA v2.2 & 33.89  \\ 
                & $\mathbf{SFMViT}$$^\ddag$ & ViT-Giant+SF-R50-NL & K400(ft:AVA v2.2) &\textbf{42.28}  \\
                \shline
            \end{tabular}
        \caption{\textbf{Comparison of Chaotic World datasets using ground truth instead of detector results.}}
        \label{table:backbone_combine}
        \end{subtable}

    \end{minipage}
    \vspace{-1mm}
    \caption{\textbf{
    Comparison with state-of-the-art models on the Chaotic World dataset.} We have bolded the best mAP.``ft" denotes fine-tuning on this dataset.$^\ddag$  denotes the model contains a multi-stream backbone. ``cap" is the abbreviation of capacity. Specifically, in Table (b), the SlowFast+ViT+Hiera model, the Hiera-Large backbone is pre-trained on K400, while SF-R50-NL and ViT-Giant are pre-trained on K400 and fine-tuned on AVA v2.2.}
\end{table*}

\begin{figure}[t]
    \centering
    \vspace{0.5mm}
    \includegraphics[width=1\linewidth]{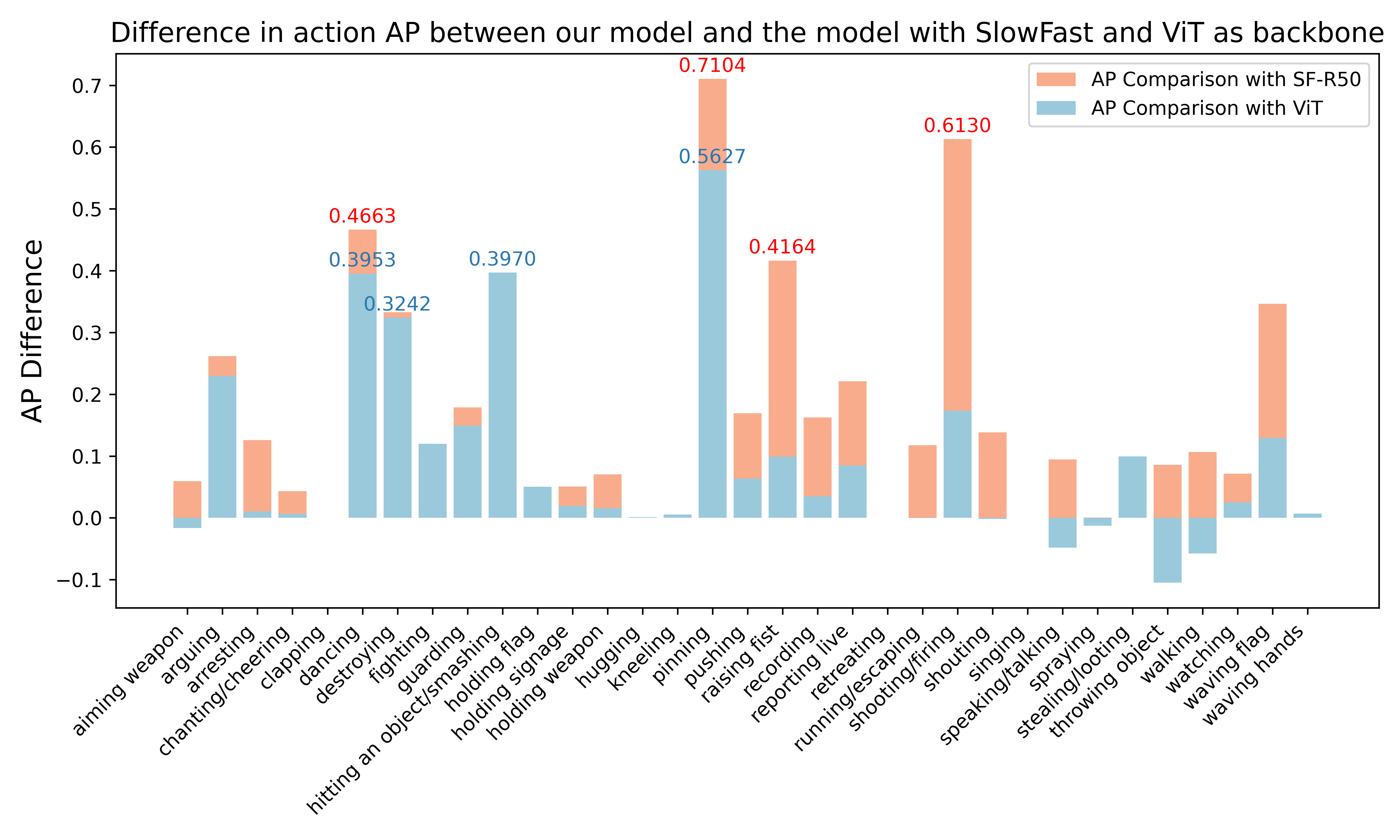}
    \vspace{-8mm}
    \caption{\textbf{Difference in action AP between our model and the model with SlowFast and ViT as backbone}. The top 5 action categories in terms of absolute value of AP change in the table are labeled with spec change values. }
    \label{fig:AP_compare_best_all}
\end{figure}

We performed an in-depth ablation study to investigate the effectiveness of our design in SFMViT. All results are reported on the Chaotic World dataset. Table \ref{tab:ablation-comparison} shows the frame-mAP comparisons under different ablation settings. Here, ``cap" stands for capacity, and the ablation study regarding capacity is conducted with SFMViT as the baseline.

\textbf{Backbone Comparison.} As shown in Table \ref{tab:ablation-comparison}, under the condition of capacity=50, SFMViT's mAP is significantly higher than that of single-stream backbone models. This illustrates that our dual-stream backbone complements each other, and the absence of either stream would result in a significant decrease in mAP. In Fig.\ref{fig:AP_compare_best_all}, we show a comparison of the action APs of the dual-stream backbone and each stream backbone (SlowFast-ResNet50 and ViT-Giant) in the Chaotic World dataset. Hereafter, we refer to the model with SlowFast-ResNet50 as the backbone as SlowFast*, and the model with ViT-Giant as the backbone as ViT*. 

SFMViT outperforms single-stream backbone models in most action categories and even surpasses all action APs of SlowFast*. This is because the dual-stream backbone combines the powerful global feature extraction capability of ViT with the excellent spatiotemporal modeling capability of SlowFast, a conclusion confirmed in \cite{dosovitskiy2020image,feichtenhofer2019slowfast}.

However, compared to ViT*, SFMViT exhibits a slight decrease in AP for five action categories such as ``throwing object" and ``walking". This is primarily because SlowFast performs poorly in these categories, leading to a decline in the performance of the fusion model. The reasons can be summarized as follows: \Rmnum{1}. These actions are not time-sensitive. Taking the ``throwing object" action as an example, the key recognition features mainly focus on the movement of the arm from backward swing to forward throwing, where the speed or accuracy at specific time points is not critical, but rather the trajectory and posture of the arm. \Rmnum{2}. SlowFast exhibits a strong bias in action judgment, resulting in a high misclassification rate. SlowFast* has a high misclassification rate for ``walking", often confusing it with categories such as ``holding signage" that have more training data instances or high temporal similarity. Although the introduction of SlowFast may cause some action APs to decrease slightly, due to SlowFast's advantages in spatiotemporal modeling and prior knowledge of spatiotemporal action localization, it can assist ViT and improve model expression capabilities in a short time.

\textbf{Comparison of backbone combination methods.} Not all advanced video feature extraction models or multi-stream model combinations can achieve the effectiveness of SFMViT. We compared various advanced video backbones such as Hiera and ViT as single-stream backbones and further explored combinations of these backbones into dual-stream or triple-stream configurations. To minimize the bias introduced by detectors on different models and datasets, we replaced detector annotations with ground truth, and conducted comparisons between SFMViT and other methods on a unified baseline, as shown in Table \ref{table:backbone_combine}. The mAP of SFMViT exceeds that of all single-stream and multi-stream backbone models, indicating that the dual-stream configuration of SFMViT is the most appropriate. Additionally, some backbone models (Hiera+ViT, SlowFast+ViT+Hiera) exhibit lower performance after fusion, suggesting that incompatible backbones may interfere with each other, further validating the effectiveness of SFMViT's dual-stream configuration.


\label{sec:capacity}
\begin{figure}[t]
    \centering
    \includegraphics[width=0.9\linewidth]{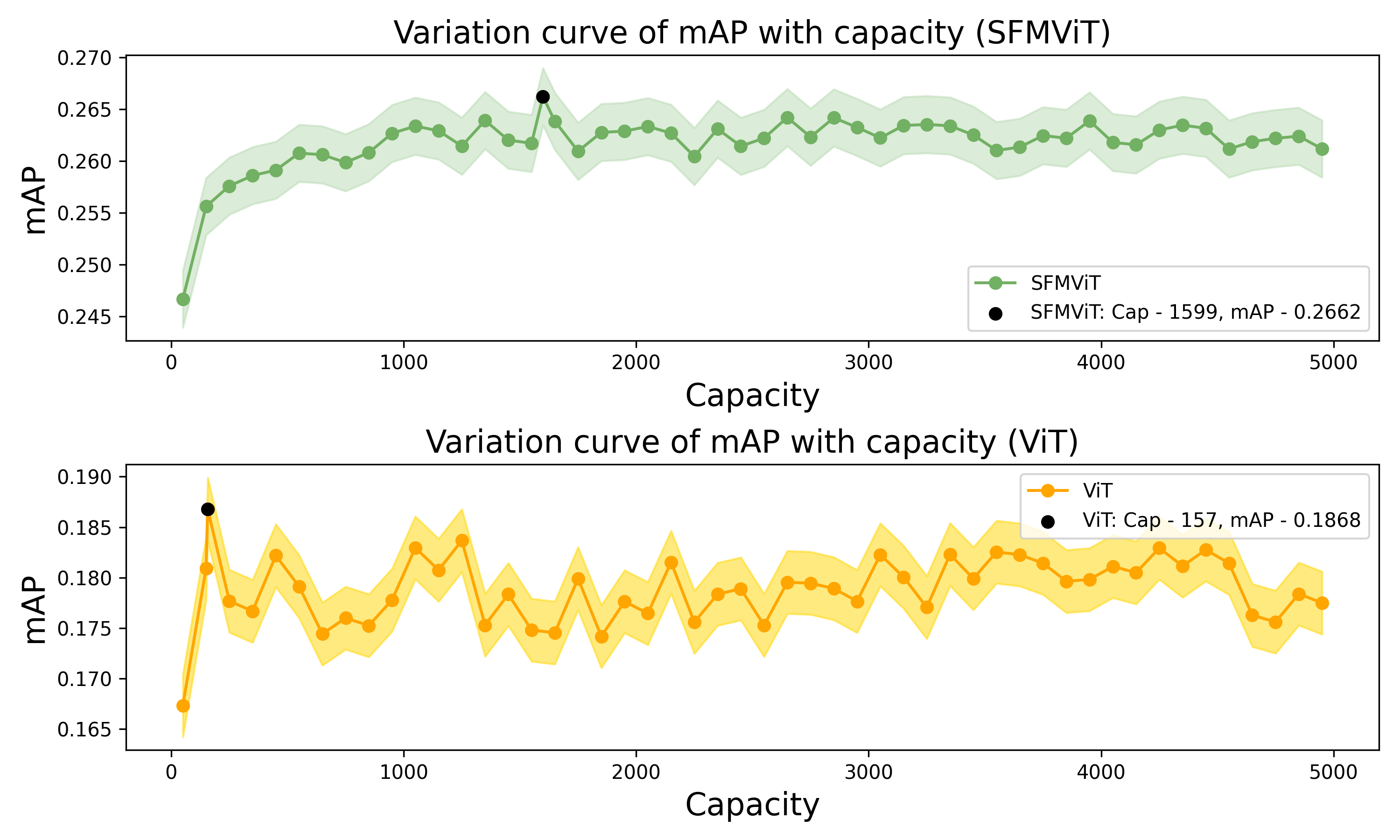}
    \caption{\textbf{Curve of mAP with capacity for SFMViT and ViT*.} The envelope outside the curve is centered on the curve, with {the curve value ± standard deviation} at that point as the upper and lower limits.}
    \label{fig:mAP_vs_Capacity}
\end{figure}

\textbf{Confidence Pruning Strategy: Capacity Comparison.} Fig. \ref{fig:mAP_vs_Capacity} shows how the model's mAP changes with capacity, with the size of the envelope reflecting the {standard deviation} of the mAP change. Capacity is the upper limit of the number of anchors predicted for each $\{img\_id, time\}$ key-value pair. As the capacity increases, mAP first rises and then slowly decreases. SFMViT obtains a maximum mAP value of 26.62\% when capacity=1599, and ViT* obtains a maximum mAP value of 18.68\% when capacity=157. If the upper limit of the number of anchors is set too small, correctly predicted anchors may be missed, thus affecting performance; if it is set too large, it will affect efficiency and consume memory. In addition, for different data and models, the mAP value is random, so it is not suitable to fix a capacity value to find the best mAP. Therefore, we can find an optimal capacity in a short time through multi-thread iteration, thereby capturing the optimal mAP value.

\vspace{-2mm}
\section{Conclusions}
\vspace{-2mm}
The Chaotic World dataset contains unique fine-grained complex actions such as destructive behaviors and lifesaving behaviors. In response to this characteristic: \Rmnum{1}. SFMViT leverages expert knowledge, pre-trained on large-scale datasets, and fine-tuned on Chaotic World. \Rmnum{2}. Its dual-stream backbone combines ViT for global features and SlowFast for spatiotemporal modeling. \Rmnum{3}. SFMViT introduces a Confidence Pruning Strategy, which can prune and select the optimal anchors, unleashing the potential of the model.

However, due to time constraints: \Rmnum{1}. Our fusion method of the dual-stream backbone is simplistic. \Rmnum{2}. Our model's performance relies heavily on the detector. In the future, we will have more time to optimize these shortcomings to fully realize the potential of our model.

\vspace{-0.3cm}
\section{Foundation}
\vspace{-2mm}
This work was supported by National Natural Science Foundation of China (No. 62203476), Natural Science Foundation of Shenzhen (No. JCYJ20230807120801002).

\vspace{-4mm}
\renewcommand{\bibfont}{\footnotesize}
\printbibliography 

@inproceedings{pan2021actor,
  title={Actor-context-actor relation network for spatio-temporal action localization},
  author={Pan, Junting and Chen, Siyu and Shou, Mike Zheng and Liu, Yu and Shao, Jing and Li, Hongsheng},
  booktitle={Proceedings of the IEEE/CVF Conference on Computer Vision and Pattern Recognition},
  pages={464--474},
  year={2021}
}

@inproceedings{wang2023videomae,
  title={Videomae v2: Scaling video masked autoencoders with dual masking},
  author={Wang, Limin and Huang, Bingkun and Zhao, Zhiyu and Tong, Zhan and He, Yinan and Wang, Yi and Wang, Yali and Qiao, Yu},
  booktitle={Proceedings of the IEEE/CVF Conference on Computer Vision and Pattern Recognition},
  pages={14549--14560},
  year={2023}
}

@inproceedings{tang2020asynchronous,
  title={Asynchronous interaction aggregation for action detection},
  author={Tang, Jiajun and Xia, Jin and Mu, Xinzhi and Pang, Bo and Lu, Cewu},
  booktitle={Computer Vision--ECCV 2020: 16th European Conference, Glasgow, UK, August 23--28, 2020, Proceedings, Part XV 16},
  pages={71--87},
  year={2020},
  organization={Springer}
}

@inproceedings{feichtenhofer2020x3d,
  title={X3d: Expanding architectures for efficient video recognition},
  author={Feichtenhofer, Christoph},
  booktitle={Proceedings of the IEEE/CVF conference on computer vision and pattern recognition},
  pages={203--213},
  year={2020}
}

@inproceedings{chen2021watch,
  title={Watch only once: An end-to-end video action detection framework},
  author={Chen, Shoufa and Sun, Peize and Xie, Enze and Ge, Chongjian and Wu, Jiannan and Ma, Lan and Shen, Jiajun and Luo, Ping},
  booktitle={Proceedings of the IEEE/CVF International Conference on Computer Vision},
  pages={8178--8187},
  year={2021}
}

@inproceedings{zhao2022tuber,
  title={Tuber: Tubelet transformer for video action detection},
  author={Zhao, Jiaojiao and Zhang, Yanyi and Li, Xinyu and Chen, Hao and Shuai, Bing and Xu, Mingze and Liu, Chunhui and Kundu, Kaustav and Xiong, Yuanjun and Modolo, Davide and others},
  booktitle={Proceedings of the IEEE/CVF Conference on Computer Vision and Pattern Recognition},
  pages={13598--13607},
  year={2022}
}

@inproceedings{wu2023stmixer,
  title={Stmixer: A one-stage sparse action detector},
  author={Wu, Tao and Cao, Mengqi and Gao, Ziteng and Wu, Gangshan and Wang, Limin},
  booktitle={Proceedings of the IEEE/CVF Conference on Computer Vision and Pattern Recognition},
  pages={14720--14729},
  year={2023}
}

@inproceedings{chen2023efficient,
  title={Efficient video action detection with token dropout and context refinement},
  author={Chen, Lei and Tong, Zhan and Song, Yibing and Wu, Gangshan and Wang, Limin},
  booktitle={Proceedings of the IEEE/CVF International Conference on Computer Vision},
  pages={10388--10399},
  year={2023}
}

@InProceedings{Ong_2023_ICCV,
author    = {Ong, Kian Eng and Ng, Xun Long and Li, Yanchao and Ai, Wenjie and Zhao, Kuangyi and Yeo, Si Yong and Liu, Jun},
title     = {Chaotic World: A Large and Challenging Benchmark for Human Behavior Understanding in Chaotic Events},
booktitle = {Proceedings of the IEEE/CVF International Conference on Computer Vision (ICCV)},
month     = {October},
year      = {2023},
pages     = {20213-20223}
}

@inproceedings{gu2018ava,
  title={Ava: A video dataset of spatio-temporally localized atomic visual actions},
  author={Gu, Chunhui and Sun, Chen and Ross, David A and Vondrick, Carl and Pantofaru, Caroline and Li, Yeqing and Vijayanarasimhan, Sudheendra and Toderici, George and Ricco, Susanna and Sukthankar, Rahul and others},
  booktitle={Proceedings of the IEEE conference on computer vision and pattern recognition},
  pages={6047--6056},
  year={2018}
}

@article{dosovitskiy2020image,
  title={An image is worth 16x16 words: Transformers for image recognition at scale},
  author={Dosovitskiy, Alexey and Beyer, Lucas and Kolesnikov, Alexander and Weissenborn, Dirk and Zhai, Xiaohua and Unterthiner, Thomas and Dehghani, Mostafa and Minderer, Matthias and Heigold, Georg and Gelly, Sylvain and others},
  journal={arXiv preprint arXiv:2010.11929},
  year={2020}
}

@inproceedings{feichtenhofer2019slowfast,
  title={Slowfast networks for video recognition},
  author={Feichtenhofer, Christoph and Fan, Haoqi and Malik, Jitendra and He, Kaiming},
  booktitle={Proceedings of the IEEE/CVF international conference on computer vision},
  pages={6202--6211},
  year={2019}
}

@inproceedings{ryali2023hiera,
  title={Hiera: A hierarchical vision transformer without the bells-and-whistles},
  author={Ryali, Chaitanya and Hu, Yuan-Ting and Bolya, Daniel and Wei, Chen and Fan, Haoqi and Huang, Po-Yao and Aggarwal, Vaibhav and Chowdhury, Arkabandhu and Poursaeed, Omid and Hoffman, Judy and others},
  booktitle={International Conference on Machine Learning},
  pages={29441--29454},
  year={2023},
  organization={PMLR}
}

@inproceedings{sun2018actor,
  title={Actor-centric relation network},
  author={Sun, Chen and Shrivastava, Abhinav and Vondrick, Carl and Murphy, Kevin and Sukthankar, Rahul and Schmid, Cordelia},
  booktitle={Proceedings of the European Conference on Computer Vision (ECCV)},
  pages={318--334},
  year={2018}
}

@inproceedings{carreira2017quo,
  title={Quo vadis, action recognition? a new model and the kinetics dataset},
  author={Carreira, Joao and Zisserman, Andrew},
  booktitle={proceedings of the IEEE Conference on Computer Vision and Pattern Recognition},
  pages={6299--6308},
  year={2017}
}

@inproceedings{xie2018rethinking,
  title={Rethinking spatiotemporal feature learning: Speed-accuracy trade-offs in video classification},
  author={Xie, Saining and Sun, Chen and Huang, Jonathan and Tu, Zhuowen and Murphy, Kevin},
  booktitle={Proceedings of the European conference on computer vision (ECCV)},
  pages={305--321},
  year={2018}
}

@article{wang2024internvideo2,
  title={InternVideo2: Scaling Video Foundation Models for Multimodal Video Understanding},
  author={Wang, Yi and Li, Kunchang and Li, Xinhao and Yu, Jiashuo and He, Yinan and Chen, Guo and Pei, Baoqi and Zheng, Rongkun and Xu, Jilan and Wang, Zun and others},
  journal={arXiv preprint arXiv:2403.15377},
  year={2024}
}

@article{kay2017kinetics,
  title={The kinetics human action video dataset},
  author={Kay, Will and Carreira, Joao and Simonyan, Karen and Zhang, Brian and Hillier, Chloe and Vijayanarasimhan, Sudheendra and Viola, Fabio and Green, Tim and Back, Trevor and Natsev, Paul and others},
  journal={arXiv preprint arXiv:1705.06950},
  year={2017}
}

@inproceedings{sigurdsson2016hollywood,
  title={Hollywood in homes: Crowdsourcing data collection for activity understanding},
  author={Sigurdsson, Gunnar A and Varol, G{\"u}l and Wang, Xiaolong and Farhadi, Ali and Laptev, Ivan and Gupta, Abhinav},
  booktitle={Computer Vision--ECCV 2016: 14th European Conference, Amsterdam, The Netherlands, October 11--14, 2016, Proceedings, Part I 14},
  pages={510--526},
  year={2016},
  organization={Springer}
}

@inproceedings{li2023unmasked,
  title={Unmasked teacher: Towards training-efficient video foundation models},
  author={Li, Kunchang and Wang, Yali and Li, Yizhuo and Wang, Yi and He, Yinan and Wang, Limin and Qiao, Yu},
  booktitle={Proceedings of the IEEE/CVF International Conference on Computer Vision},
  pages={19948--19960},
  year={2023}
}

@ARTICLE{8764496,
  author={Zhang, Shifeng and Xie, Yiliang and Wan, Jun and Xia, Hansheng and Li, Stan Z. and Guo, Guodong},
  journal={IEEE Transactions on Multimedia}, 
  title={WiderPerson: A Diverse Dataset for Dense Pedestrian Detection in the Wild}, 
  year={2020},
  volume={22},
  number={2},
  pages={380-393},
  doi={10.1109/TMM.2019.2929005}}

@article{wang2024yolov9,
  title={YOLOv9: Learning What You Want to Learn Using Programmable Gradient Information},
  author={Wang, Chien-Yao and Yeh, I-Hau and Liao, Hong-Yuan Mark},
  journal={arXiv preprint arXiv:2402.13616},
  year={2024}
}

@article{liu2022generalized,
  title={Generalized Pose Decoupled Network for Unsupervised 3d Skeleton Sequence-based Action Representation Learning},
  author={Liu, Mengyuan and Meng, Fanyang and Liang, Yongsheng},
  journal={Cyborg and Bionic Systems},
  volume={2022},
  pages={0002},
  year={2022}
}

\end{document}